\documentclass{article} 
\usepackage{iclr2017_conference,times}
\usepackage{hyperref}
\usepackage{url}
\usepackage{graphicx}
\usepackage{array}
\usepackage{subfigure}
\usepackage{graphicx}
\usepackage{tablefootnote}
\usepackage{blindtext}
\usepackage[english]{babel}
\usepackage{amsmath}
\usepackage[normalem]{ulem}
\usepackage{footnote}
\usepackage{latexsym}
\usepackage{multirow}
\usepackage[linesnumbered,ruled]{algorithm2e}

\newcommand{\cev}[1]{\reflectbox{\ensuremath{\vec{\reflectbox{\ensuremath{#1}}}}}}

\title{S-Net: From Answer Extraction to Answer Generation for Machine Reading Comprehension}

\author{Chuanqi Tan$^\dag$\thanks{\;Contribution during internship at Microsoft Research.} , Furu Wei$^\ddag$, Nan Yang$^\ddag$, Bowen Du$^\dag$, Weifeng Lv$^\dag$, Ming Zhou$^\ddag$ \\
	$^\dag$ State Key Laboratory of Software Development Environment, Beihang University, Beijing, China \\ 
	$^\ddag$ Microsoft Research, Beijing, China \\
	{\tt tanchuanqi@nlsde.buaa.edu.cn} \hspace{0.5cm} {\tt \{dubowen,lwf\}@buaa.edu.cn} \\
	\texttt{\{fuwei,nanya,mingzhou\}@microsoft.com} \\
}

\begin{document}

	\maketitle
		\begin{abstract}	
			In this paper, we present a novel approach to machine reading comprehension for the MS-MARCO dataset. Unlike the SQuAD dataset that aims to answer a question with exact text spans in a passage, the MS-MARCO dataset defines the task as answering a question from multiple passages and the words in the answer are not necessary in the passages. We therefore develop an extraction-then-synthesis framework to synthesize answers from extraction results. Specifically, the answer extraction model is first employed to predict the most important sub-spans from the passage as evidence, and the answer synthesis model takes the evidence as additional features along with the question and passage to further elaborate the final answers. We build the answer extraction model with state-of-the-art neural networks for single passage reading comprehension, and propose an additional task of passage ranking to help answer extraction in multiple passages. 
			The answer synthesis model is based on the sequence-to-sequence neural networks with extracted evidences as features. Experiments show that our extraction-then-synthesis method outperforms state-of-the-art methods.
			
		\end{abstract}
		
		\section{Introduction}
		
		Machine reading comprehension~\citep{D16-1264,marco}, which attempts to enable machines to answer questions after reading a passage or a set of passages, attracts great attentions from both research and industry communities in recent years. The release of the Stanford Question Answering Dataset (SQuAD)~\citep{D16-1264} and the Microsoft MAchine Reading COmprehension Dataset (MS-MARCO)~\citep{marco} provides the large-scale manually created datasets for model training and testing of machine learning (especially deep learning) algorithms for this task. There are two main differences in existing machine reading comprehension datasets. First, the SQuAD dataset constrains the answer to be an exact sub-span in the passage, while words in the answer are not necessary in the passages in the MS-MARCO dataset. Second, the SQuAD dataset only has one passage for a question, while the MS-MARCO dataset contains multiple passages.	
		
		Existing methods for the MS-MARCO dataset usually follow the extraction based approach for single passage in the SQuAD dataset. It formulates the task as predicting the start and end positions of the answer in the passage. However, as defined in the MS-MARCO dataset, the answer may come from multiple spans, and the system needs to elaborate the answer using words in the passages and words from the questions as well as words that cannot be found in the passages or questions. 
		
		Table \ref{example} shows several examples from the MS-MARCO dataset. Except in the first example the answer is an exact text span in the passage, in other examples the answers need to be synthesized or generated from the question and passage. In the second example the answer consists of multiple text spans (hereafter evidence snippets) from the passage. In the third example, the answer contains words from the question. In the fourth example, the answer has words that cannot be found in the passages or question. In the last example, all words are not in the passages or questions.
		
\begin{figure}[h]
			\begin{center}
				\includegraphics[width=3.3in]{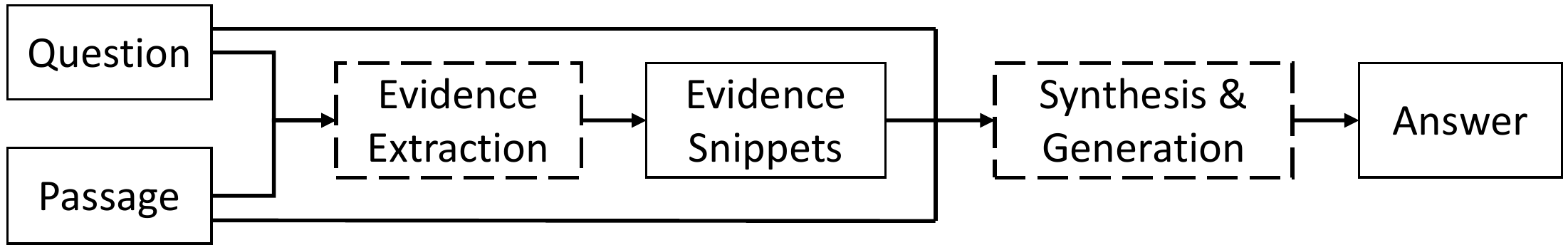}
			\end{center}
			\caption{Overview of S-Net. It first extracts evidence snippets by matching the question and passage, and then generates the answer by synthesizing the question, passage, and evidence snippets.}
			\label{overview}
		\end{figure}

		In this paper, we present an extraction-then-synthesis framework for machine reading comprehension shown in Figure~\ref{overview}, in which the answer is synthesized from the extraction results. We build an evidence extraction model to predict the most important sub-spans from the passages as evidence, and then develop an answer synthesis model which takes the evidence as additional features along with the question and passage to further elaborate the final answers.
		
		\begin{table*}[h]
			\centering
			\begin{tabular}{c p{13cm}}
				\hline
				\multicolumn{2}{l}{\textbf{The answer is an exact text span in the passage.}} \\
				& $Q$: how tall is jack griffo \\
				& $P$: Jack Griffo Height : \textbf{5'6 (167.64 cm)}. Standing at a height of 5 feet, 6 inches tall Jack Griffo is taller than 11.9\% of all men, as reflected by the figure's fill height \%.Conversely, at this height Jack Griffo is not as tall as 88.1\% of all men. \\
				& $A$: 5'6 (167.64 cm) \\
				\hline
				\multicolumn{2}{l}{\textbf{All words in the answer are in the passage but from multiple text spans.}} \\
				& $Q$: who did odysseus see in the underworld \\					
				& $P_1$: The souls that Odysseus saw in the Underworld On seeing \textbf{Achilles}' soul, said Odysseus: Achilles, the most fortunate man that ever was or will be honored as though you were a god and now you are a mighty prince among the dead. \\
				& $P_2$: Odysseus talked to his mother Anticlea, who died of grief when he did not return home after the Trojan War. Odysseus was also surprised to see \textbf{Elphenor}, the youngest member of his crew, in the Underworld. \\
				& $A$: Elphenor and Achilles.	 \\			
				\hline
				
				\multicolumn{2}{l}{\textbf{All words in the answer are in the passage and question.} }\\ 
				& $Q$: what do \textbf{producers need to make food} \\
				& $P$: Plants are producers. Producers are living things that can make their own food using \textbf{air, light, soil, and water}. Plants use a process called photosynthesis to make food. \\
				& $A$: Producers need air, light, soil, and water to make food. \\
				\hline
				\multicolumn{2}{l}{\textbf{Part of words in the answer are not found in the passage or question.}} \\
				& $Q$: why conversion observed in body \\
				& $P$: Conversion disorder \textbf{symptoms} may appear suddenly after a stressful event or trauma, whether physical or psychological. Signs and symptoms that affect movement function may include: 1 Weakness or paralysis. 2 Abnormal movement, such as tremors or difficulty walking. 3 Loss of balance. \\
				& $A$: Due to symptoms in the body\\
				\hline
				
				\multicolumn{2}{l}{\textbf{All Words in the answer are not found in the passages or question.}} \\
				& $Q$: is there an age limit for learning speech \\
				& $P$: Age is not a detriment to language learning, and by all accounts, learning a second (or third etc) language actually keeps the older language learners mind active. People of all ages can benefit from learning languages. \\
				& $A$: No \\
				\hline
			\end{tabular}
			\caption{Representative examples for different kinds of answer according to the necessary of synthesis in the MS-MARCO dataset. $Q$, $P$, and $A$ represent question, passage, and answer, respectively. The text in bold is the evidence in the passage or the word in the question that matches the answer. }
			\label{example}
		\end{table*}
		
		Specifically, we develop the answer extraction model with state-of-the-art attention based neural networks which predict the start and end positions of evidence snippets. As multiple passages are provided for each question in the MS-MARCO dataset, we propose incorporating passage ranking as an additional task to improve the results of evidence extraction under a multi-task learning framework. We use the bidirectional recurrent neural networks (RNN) for the word-level representation, and then apply the attention mechanism \citep{RocktaschelGHKB15} to incorporate matching information from question to passage at the word level. Next, we predict start and end positions of the evidence snippet by pointer networks~\citep{NIPS2015_5866}. Moreover, we aggregate the word-level matching information of each passage using the attention pooling, and use the passage-level representation to rank all candidate passages as an additional task. For the answer synthesis, we apply the sequence-to-sequence model to synthesize the final answer based on the extracted evidence. The question and passage are encoded by a bi-directional RNN in which the start and end positions of extracted snippet are labeled as features. We combine the question and passage information in the encoding part to initialize the attention-equipped decoder to generate the answer.
		
		We conduct experiments on the MS-MARCO dataset. The results show our extraction-then-synthesis framework outperforms our baselines and all other existing methods in terms of ROUGE-L and BLEU-1.
		
		Our contributions can be summarized as follows:
		\begin{itemize}
			\item	We propose an extraction-then-synthesis framework for machine reading comprehension in which words in answer are not necessary in the passages.
			\item	We incorporate passage ranking to pure answer span prediction, which improves the extraction result in the multiple passages reading comprehension.
			\item	We develop an answer synthesis model that applies the sequence-to-sequence model to generate the answer with extracted evidences as features, which outperforms pure answer extraction methods and all other existing methods on the MS-MARCO dataset.
		\end{itemize}
		
		\section{Related Work}
		Benchmark datasets play an important role in recent progress in reading comprehension and question answering research. \citet{Richardson2013} release MCTest whose goal is to select the best answer from four options given the question and the passage. CNN/Daily-Mail~\citep{Hermann2015} and CBT~\citep{Hill2016} are the cloze-style datasets in which the goal is to predict the missing word (often a named entity) in a passage. Different from above datasets, the SQuAD dataset \citep{D16-1264} whose answer can be much longer phrase is more challenging. The answer in SQuAD is a segment of text, or span, from the corresponding reading passage. Similar to the SQuAD, MS-MARCO~\citep{marco} is the reading comprehension dataset which aims to answer the question given a set of passages. The answer in MS-MARCO is generated by human after reading all related passages and not necessarily sub-spans of the passages. 
		
		To the best of our knowledge, the existing works on the MS-MARCO dataset follow their methods on the SQuAD. 
		\citet{wang2016machine} combine match-LSTM and pointer networks to produce the boundary of the answer. \citet{xiong2016dynamic} and \citet{seo2016bidirectional} employ variant co-attention mechanism to match the question and passage mutually. \citet{xiong2016dynamic} propose a dynamic pointer network to iteratively infer the answer. \citet{rnet} apply an additional gate to the attention-based recurrent networks and propose a self-matching mechanism for aggregating evidence from the whole passage, which achieves the state-of-the-art result on SQuAD dataset. Other works which only focus on the SQuAD dataset may also be applied on the MS-MARCO dataset \citep{yu2016end,lee2016learning,yang2016}.
		
		The sequence-to-sequence model is widely-used in many tasks such as machine translation \citep{D15-1166}, parsing \citep{vinyals2015grammar}, response generation \citep{P16-1154}, and summarization generation \citep{zhou_acl17_selective}. We use it to generate the synthetic answer with the start and end positions of the evidence snippet as features.
		
		\section{Our Approach}
		
		Following the overview in Figure~\ref{overview}, our approach consists of two parts as evidence extraction\footnote{In our model, we use ``evidence extraction'' to represent the pure ``answer extraction'' in previous work.} and answer synthesis. The two parts are trained in two stages. The evidence extraction part aims to extract evidence snippets related to the question and passage. The answer synthesis part aims to generate the answer based on the extracted evidence snippets. We propose a multi-task learning framework for the evidence extraction shown in Figure~\ref{r-network_overview}, and use the sequence-to-sequence model with additional features of the start and end positions of the evidence snippet for the answer synthesis shown in Figure~\ref{s2s-network_overview}.
		
		\subsection{Gated Recurrent Unit}
		
		We use Gated Recurrent Unit (GRU) \citep{cho2014} instead of basic RNN. Equation \ref{eq:gru} describes the mathematical model of the GRU. $r_t$ and $z_t$ are the gates and $h_t$ is the hidden state.

		\begin{align}
		z_t &= \sigma(W_{hz} h_{t-1} + W_{xz} x_t + b_z)\nonumber\\
		r_t &= \sigma(W_{hr} h_{t-1} + W_{xr} x_t + b_r)\nonumber\\
		\hat{h_t} &= \Phi(W_h (r_t \odot h_{t-1}) + W_x x_t + b)\nonumber\\
		h_t &= (1-z_t)\odot h_{t-1} + z_t \odot \hat{h_t} 
		\label{eq:gru}
		\end{align}

		\subsection{Evidence Extraction}
		
\begin{figure}
				\begin{center}
					\includegraphics[width=3in]{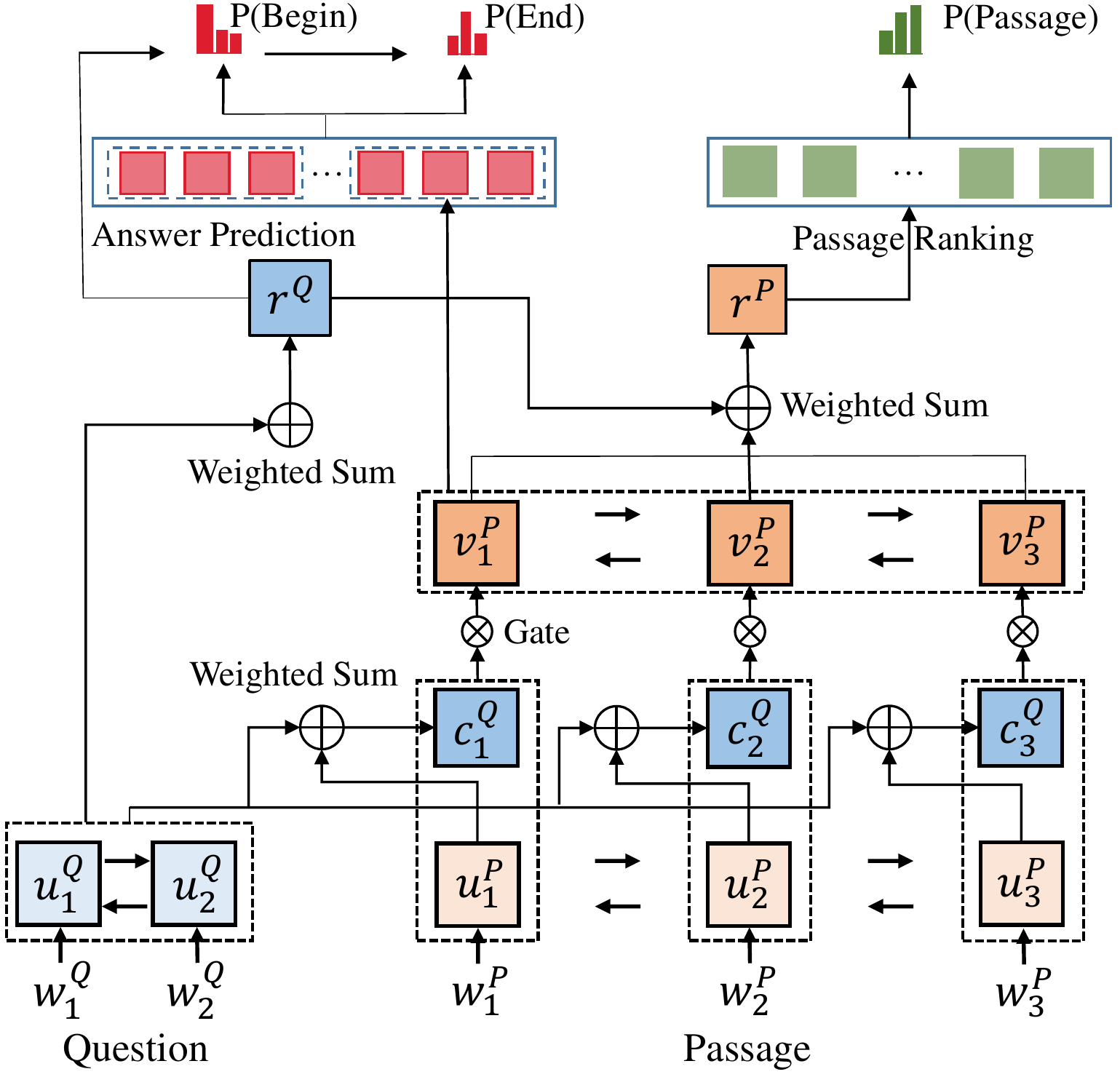}
				\end{center}
				\caption{Evidence Extraction Model}
				\label{r-network_overview}
			\end{figure}

		We propose a multi-task learning framework for evidence extraction. Unlike the SQuAD dataset, which only has one passage given a question, there are several related passages for each question in the MS-MARCO dataset. In addition to annotating the answer, MS-MARCO also annotates which passage is correct. To this end, we propose improving text span prediction with passage ranking. Specifically, as shown in Figure~\ref{r-network_overview}, in addition to predicting a text span, we apply another task to rank candidate passages with the passage-level representation. 		
		
		\subsubsection{Evidence Snippet Prediction}
		
		Consider a question Q = $\{w_t^Q\}_{t=1}^m$ and a passage P = $\{w_t^P\}_{t=1}^n$, we first convert the words to their respective word-level embeddings and character-level embeddings. The character-level embeddings are generated by taking the final hidden states of a bi-directional GRU applied to embeddings of characters in the token. We then use a bi-directional GRU to produce new representation $u^Q_1, \dots, u^Q_m$ and $u^P_1, \dots, u^P_n$ of all words in the question and passage respectively:
		\begin{align}
		u_t^Q = \mathrm{BiGRU}_Q(u_{t - 1}^Q, [e_t^Q,char_t^Q]) \nonumber\\
		u_t^P = \mathrm{BiGRU}_P(u_{t - 1}^P, [e_t^P,char_t^P])
		\end{align}
		Given question and passage representation $\{u_t^Q\}_{t=1}^m$ and $\{u_t^P\}_{t=1}^n$, \citet{RocktaschelGHKB15} propose generating sentence-pair representation $\{v_t^P\}_{t=1}^n$ via soft-alignment of words in the question and passage as follows:
		\begin{equation}
		v_t^P = \mathrm{GRU} (v_{t-1}^P, c^Q_t)
		\end{equation}
		where $c^Q_t=att(u^Q, [u_t^P, v_{t-1}^P])$ is an attention-pooling vector of the whole question ($u^Q$):
		\begin{align}
		s_j^t &= \mathrm{v}^\mathrm{T}\mathrm{tanh}(W_u^Q u_j^Q + W_u^P u_t^P) \nonumber \\
		a_i^t &= \mathrm{exp}(s_i^t) / \Sigma_{j=1}^m \mathrm{exp}(s_j^t) \nonumber \\
		c^Q_t &= \Sigma_{i=1}^m a_i^t u_i^Q
		\end{align}	
		\citet{WangJ16} introduce match-LSTM, which takes $u_j^P$ as an additional input into the recurrent network. \citet{rnet} propose adding gate to the input ($[u_t^P, c^Q_t]$) of RNN to determine the importance of passage parts. 
		\begin{align}
		&g_t = \mathrm{sigmoid}(W_g [u_t^P, c^Q_t]) \nonumber \\
		&[u_t^P, c^Q_t]^* = g_t\odot[u_t^P, c^Q_t] \nonumber \\
		&v_t^P = \mathrm{GRU} (v_{t-1}^P, [u_t^P, c^Q_t]^*)
		\label{vtp}
		\end{align}
		
		We use pointer networks~\citep{NIPS2015_5866} to predict the position of evidence snippets. 	
		Following the previous work \citep{wang2016machine}, we concatenate all passages to predict one span for the evidence snippet prediction. Given the representation $\{v_t^P\}_{t=1}^N$ where $N$ is the sum of the length of all passages, the attention mechanism is utilized as a pointer to select the start position ($p^1$) and end position ($p^2$), which can be formulated as follows:
		\begin{align}
		s_j^t &= \mathrm{v}^\mathrm{T}\mathrm{tanh}(W_h^{P} v_j^P + W_{h}^{a} h_{t-1}^a) \nonumber \\
		a_i^t &= \mathrm{exp}(s_i^t) / \Sigma_{j=1}^N \mathrm{exp}(s_j^t) \nonumber \\
		p^t &= \mathrm{argmax}(a_1^t, \dots, a_N^t)
		\end{align}
		Here $h_{t-1}^a$ represents the last hidden state of the answer recurrent network (pointer network). The input of the answer recurrent network is the attention-pooling vector based on current predicted probability $a^t$:
		\begin{align}
		c_t &= \Sigma_{i=1}^N a_i^t v_i^P \nonumber \\ 
		h_t^a &= \mathrm{GRU}(h_{t-1}^a, c_t)
		\end{align}
		When predicting the start position, $h_{t-1}^a$ represents the initial hidden state of the answer recurrent network. We utilize the question vector $r^Q$ as the initial state of the answer recurrent network. $r^Q = att(u^Q, v^Q_r)$ is an attention-pooling vector of the question based on the parameter $v^Q_r$:
		\begin{align}
		s_j &= \mathrm{v}^\mathrm{T}\mathrm{tanh}(W_u^{Q} u_j^Q + W_{v}^{Q} v_r^Q) \nonumber \\
		a_i &= \mathrm{exp}(s_i) / \Sigma_{j=1}^m \mathrm{exp}(s_j) \nonumber \\
		r^Q &= \Sigma_{i=1}^m a_i u_i^Q
		\end{align}
		
		For this part, the objective function is to minimize the following cross entropy:
		\begin{equation}
		\mathcal{L}_{AP} = -\Sigma_{t=1}^{2}\Sigma_{i=1}^{N}[y^t_i\log a^t_i + (1-y^t_i)\log (1-a^t_i)]
		\end{equation}
		where $y^t_i \in \{0,1\}$ denotes a label. $y^t_i=1$ means $i$ is a correct position, otherwise $y^t_i=0$.
		
		\subsubsection{Passage Ranking}
		
		In this part, we match the question and each passage from word level to passage level. Firstly, we use the question representation $r^Q$ to attend words in each passage to obtain the passage representation $r^P$ where $r^P = att(v^P, r^Q)$. 
		\begin{align}
		s_j &= \mathrm{v}^\mathrm{T}\mathrm{tanh}(W_v^{P} v_j^P + W_{v}^{Q} r^Q) \nonumber \\
		a_i &= \mathrm{exp}(s_i) / \Sigma_{j=1}^n \mathrm{exp}(s_j) \nonumber \\
		r^P &= \Sigma_{i=1}^n a_i v_i^P
		\end{align}
		Next, the question representation $r^Q$ and the passage representation $r^P$ are combined to pass two fully connected layers for a matching score,
		\begin{equation}
		g = v_g^{\mathrm{T}}(\mathrm{tanh}(W_g[r^Q,r^P]))
		\end{equation}
		For one question, each candidate passage $P_i$ has a matching score $g_i$. We normalize their scores and optimize following objective function:
		\begin{gather}
		\hat{g}_i = \mathrm{exp}(g_i) / \Sigma_{j=1}^k \mathrm{exp}(g_j) \nonumber \\
		\mathcal{L}_{PR} = -\sum_{i=1}^{k}[y_i\log \hat{g}_i + (1-y_i)\log (1-\hat{g}_i)]
		\end{gather}
		where $k$ is the number of passages. $y_i \in \{0,1\}$ denotes a label. $y_i=1$ means $P_i$ is the correct passage, otherwise $y_i=0$.
		
		\subsubsection{Joint Learning}
		
		The evident extraction part is trained by minimizing joint objective functions:
		\begin{equation}
		\mathcal{L}_{E} = r \mathcal{L}_{AP} + (1-r) \mathcal{L}_{PR}
		\end{equation}
		where $r$ is the hyper-parameter for weights of two loss functions.
		
		\subsection{Answer Synthesis}	
		
\begin{figure}
				\begin{center}
					\includegraphics[width=3.28in]{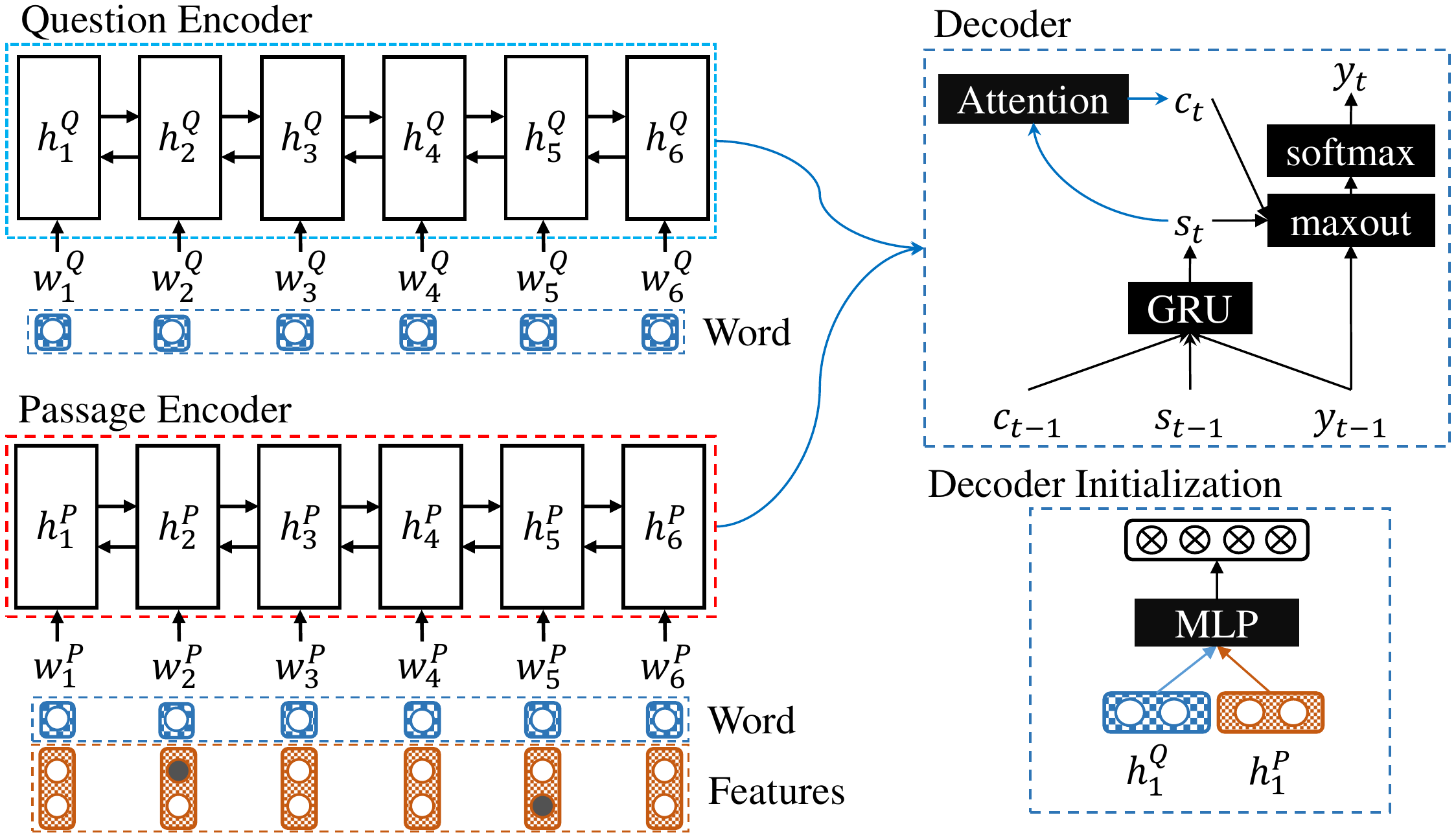}
				\end{center}
				\caption{Answer Synthesis Model}
				\label{s2s-network_overview}
			\end{figure}

		As shown in Figure~\ref{s2s-network_overview}, we use the sequence-to-sequence model to synthesize the answer with the extracted evidences as features. We first produce the representation $h_{t}^P$ and $h_{t}^Q$ of all words in the passage and question respectively. When producing the answer representation, we combine the basic word embedding $e_t^p$ with additional features $f_t^s$ and $f_t^e$ to indicate the start and end positions of the evidence snippet respectively predicted by evidence extraction model. $f_t^s =1$ and $f_t^e =1$ mean the position $t$ is the start and end of the evidence span, respectively.
		\begin{align}
		&h_{t}^P =\mathrm{BiGRU}(h_{t-1}^P, [e_t^p,f_t^s,f_t^e]) \nonumber \\
		&h_{t}^Q = \mathrm{BiGRU}(h_{t-1}^Q,e_t^Q)
		\end{align}
		
		On top of the encoder, we use GRU with attention as the decoder to produce the answer. At each decoding time step $t$, the GRU reads the previous word embedding $ w_{t-1} $ and previous context vector $ c_{t-1} $ as inputs to compute the new hidden state $ d_{t} $. To initialize the GRU hidden state, we use a linear layer with the last backward encoder hidden state $ \cev{h}_{1}^P $ and $ \cev{h}_{1}^Q $ as input:
		\begin{align}
		d_{t} &= \text{GRU}(w_{t-1}, c_{t-1}, d_{t-1}) \nonumber \\
		d_{0} &= \tanh (W_{d}[\cev{h}_{1}^P,\cev{h}_{1}^Q] + b)
		\end{align}
		where $ W_{d} $ is the weight matrix and $ b $ is the bias vector.
		
		The context vector $ c_{t} $ for current time step $ t $ is computed through the concatenate attention mechanism \citep{D15-1166}, which matches the current decoder state $ d_{t} $ with each encoder hidden state $ h_{t} $ to get the weighted sum representation. Here $h_{i}$ consists of the passage representation $h_{t}^P$ and the question representation $h_{t}^Q$. 
		\begin{align}
		s^t_j &= v_{a}^{\mathrm{T}}\tanh(W_{a}d_{t-1} + U_{a}h_{j}) \nonumber \\
		a^t_i &= \mathrm{exp}(s_i^t) / \Sigma_{j=1}^n \mathrm{exp}(s_j^t) \nonumber \\
		c_{t} &= \Sigma_{i = 1}^{n} a^t_ih_{i}
		\end{align}
		We then combine the previous word embedding $ w_{t-1} $, the current context vector $ c_{t} $, and the decoder state $ d_{t} $ to construct the readout state $ r_{t} $.
		The readout state is then passed through a maxout hidden layer \citep{Goodfellow13maxoutnetworks} to predict the next word with a softmax layer over the decoder vocabulary.
		\begin{align}
		r_{t} &= W_{r}w_{t-1} + U_{r}c_{t} + V_{r}d_{t} \nonumber \\
		\label{eq:maxout}m_{t} &= [\max\{r_{t, 2j-1}, r_{t, 2j}\}]^{\mathrm{T}} \nonumber \\
		p(y_{t} &\vert y_{1}, \dots, y_{t-1}) = \text{softmax}(W_{o}m_{t})
		\end{align}	
		where $ W_{a} $, $ U_{a} $, $ W_{r} $, $ U_{r} $, $ V_{r} $ and $ W_{o} $ are parameters to be learned.
		Readout state $ r_{t} $ is a $ 2d $-dimensional vector, and the maxout layer (Equation \ref{eq:maxout}) picks the max value for every two numbers in $ r_{t} $ and produces a d-dimensional vector $ m_{t} $.
		
		Our goal is to maximize the output probability given the input sentence. Therefore, we optimize the negative log-likelihood loss function:
		\begin{equation}
		\mathcal{L}_{S}= - \frac{1}{\vert \mathcal{D} \vert} \Sigma_{(X, Y) \in \mathcal{D}} \log p(Y|X)
		\end{equation}
		where $\mathcal{D}$ is the set of data. $X$ represents the question and passage including evidence snippets, and $Y$ represents the answer.
		
		\section{Experiment}
		
		We conduct our experiments on the MS-MARCO dataset \citep{marco}. We compare our extraction-then-synthesis framework with pure extraction model and other baseline methods on the leaderboard of MS-MARCO. Experimental results show that our model achieves better results in official evaluation metrics. We also conduct ablation tests to verify our method, and compare our framework with the end-to-end generation framework.
		
		\subsection{Dataset and Evaluation Metrics}
		
		For the MS-MARCO dataset, the questions are user queries issued to the Bing search engine and the context passages are from real web documents. The data has been split into a training set (82,326 pairs), a development set (10,047 pairs) and a test set (9,650 pairs). 
		
		The answers are human-generated and not necessarily sub-spans of the passages so that the metrics in the official tool of MS-MARCO evaluation are BLEU \citep{papineni2002BLEU} and ROUGE-L \citep{lin2004rouge}. In the official evaluation tool, the ROUGE-L is calculated by averaging the score per question, however, the BLEU is normalized with all questions. We hold that the answer should be evaluated case-by-case in the reading comprehension task. Therefore, we mainly focus on the result in the ROUGE-L.
		
		\subsection{Implementation Details}
		
		\subsubsection{Training}
		
		The evidence extraction and the answer synthesis are trained in two stages.
		
		For evidence extraction, since the answers are not necessarily sub-spans of the passages, we choose the span with the highest ROUGE-L score with the reference answer as the gold span in the training. Moreover, we only use the data whose ROUGE-L score of chosen text span is higher than 0.7, therefore we only use 71,417 training pairs in our experiments.
		
		For answer synthesis, the training data consists of two parts. First, for all passages in the training data, we choose the best span with highest ROUGE-L score as the evidence, and use the corresponding reference answer as the output. We only use the data whose ROUGE-L score of chosen evidence snippet is higher than 0.5. Second, we apply our evidence extraction model to all training data to obtain the extracted span. Then we treat the passage to which this span belongs as the input. 
		
		\subsubsection{Parameter}
		
		For answer extraction, we use 300-dimensional uncased pre-trained \textit{GloVe} embeddings~\citep{Pennington}\footnote{\url{http://nlp.stanford.edu/data/glove.6B.zip}.} for both question and passage without update during training. We use zero vectors to represent all out-of-vocabulary words. Hidden vector length is set to 150 for all layers. We also apply dropout \citep{Srivastava} between layers, with dropout rate 0.1. The weight $r$ is set to 0.8.
		
		For answer synthesis, we use an identical vocabulary set for the input and output collected from the training data. We set the vocabulary size to 30,000 according to the frequency and the other words are set to $<$unk$>$. All word embeddings are updated during the training. We set the word embedding size to 300, set the feature embedding size of start and end positions of the extracted snippet to 50, and set all GRU hidden state sizes to 150.
		
		The model is optimized using AdaDelta \citep{adadelta} with initial learning rate of 1.0. All hyper-parameters are selected on the MS-MARCO development set.
		
		\subsubsection{Decoding}
		When decoding, we first run our extraction model to obtain the extracted span, and run our synthesis model with the extracted result and the passage that contains this span. We use the beam search with beam size of 12 to generate the sequence. After the sequence-to-sequence model, we post-process the sequence with following rules:
		\begin{itemize}
			\item We only keep once if the sequence-to-sequence model generates duplicated words or phrases.
			\item For all ``$<$unk$>$'' and the word as well as phrase which are not existed in the extracted answer, we try to refine it by finding a word or phrase with the same adjacent words in the extracted span and passage. 
			\item If the generated answer only contains a single word ``$<$unk$>$'', we use the extracted span as the final answer.
		\end{itemize}
		
		\subsection{Baseline Methods}
		We conduct experiments with following settings:
		
		\textbf{S-Net (Extraction)}: the model that only has the evidence extraction part.
		
		\textbf{S-Net}: the model that consists of the evidence extraction part and the answer synthesis part.
		
		We implement two state-of-the-art baselines on reading comprehension, namely BiDAF \citep{seo2016bidirectional} and Prediction \citep{wang2016machine}, to extract text spans as evidence snippets. Moreover, we implement a baseline that only has the evidence extraction part without the passage ranking. Then we apply the answer synthesis part on top of their results. We also compare with other methods on the MS-MARCO leaderboard, including FastQAExt \citep{fastqa}, ReasoNet \citep{reasonet2016}, and R-Net \citep{rnet}.
		
		\subsection{Result}	
		
		Table \ref{marco_test_result} shows the results on the MS-MARCO test data\footnote{Baseline results are extracted from MS-MARCO leaderboard \url{http://www.msmarco.org/leaders.aspx} on Sept. 10, 2017.}. Our extraction model achieves 41.45 and 44.08 in terms of ROUGE-L and BLEU-1, respectively. Next we train the model 30 times with the same setting, and select models using a greedy search\footnote{We search models from high to low by their performances on the development set. We keep the model if adding it improves the result, otherwise discard.}. We sum the probability at each position of each single model to decide the ensemble result. Finally we select 13 models for ensemble, which achieves 42.92 and 44.97 in terms of ROUGE-L and BLEU-1, respectively, which achieves the state-of-the-art results of the extraction model. Then we test our synthesis model based on the extracted evidence. Our synthesis model achieves 3.78\% and 3.73\% improvement on the single model and ensemble model in terms of ROUGE-L, respectively. Our best result achieves 46.65 in terms of ROUGE-L and 44.78 in terms of BLEU-1, which outperforms all existing methods with a large margin and are very close to human performance. Moreover, we observe that our method only achieves significant improvement in terms of ROUGE-L compared with our baseline. The reason is that our synthesis model works better when the answer is short, which almost has no effect on BLEU as it is normalized with all questions. 
		
		\begin{table}
			\begin{center}
				\begin{tabular}{lcc}
					\hline
					\bf Method &\bf ROUGE-L & \bf BLEU-1
					\\ \hline
					FastQAExt & 33.67 & 33.93 \\
					Prediction & 37.33 & 40.72 \\
					ReasoNet & 38.81 & 39.86 \\
					R-Net & 42.89 & 42.22 \\
					\hline
					S-Net (Extraction) & 41.45 & 44.08 \\
					S-Net (Extraction, Ensemble) & 42.92 & 44.97 \\
					S-Net & 45.23 & 43.78 \\
					\bf	S-Net* & \bf 46.65 & \bf 44.78 \\
					\hline 
					Human Performance & 47 & 46 \\
					\hline 
				\end{tabular}
			\end{center}
			\caption{The performance on the MS-MARCO test set. *Using the ensemble result of extraction models as the input of the synthesis model.}
			\label{marco_test_result}
		\end{table}
		
		\begin{table}
			\small
			\begin{center}
				\begin{tabular}{lcc}
					\hline
					\multirow{2}{*}{\bf Method} & \bf 	\multirow{2}{*}{\bf Extraction} & \bf Extraction \\
					& & \bf +Synthesis \\
					\hline
					FastQAExt & 33.7 & - \\
					BiDAF & 34.89 & 38.73 \\
					Prediction & 37.54$^+$ & 41.55 \\
					\hline
					S-Net (w/o Passage Ranking) & 39.62 & 43.26 \\			
					S-Net & 42.23 & 45.95 \\
					\bf S-Net* & \bf 44.11 & \bf 47.76 \\
					\hline
				\end{tabular}
			\end{center}
			\caption{The performance on the MS-MARCO development set in terms of ROUGE-L. *Using the ensemble result of extraction models as the input of the synthesis model. $^+$\citet{wang2016machine} report their Prediction with 37.3.}
			\label{marco_dev_result}
		\end{table}
		
		Since answers on the test set are not published, we analyze our model on the development set. Table \ref{marco_dev_result} shows results on the development set in terms of ROUGE-L. As we can see, our method outperforms the baseline and several strong state-of-the-art systems. For the evidence extraction part, our proposed multi-task learning framework achieves 42.23 and 44.11 for the single and ensemble model in terms of ROUGE-L. For the answer synthesis, the single and ensemble models improve 3.72\% and 3.65\% respectively in terms of ROUGE-L. We observe the consistent improvement when applying our answer synthesis model to other answer span prediction models, such as BiDAF and Prediction. 
		
		\subsection{Discussion}
		
		\subsubsection{Ablation Test on Passage Ranking}
		
		We analyze the result of incorporating passage ranking as an additional task. We compare our multi-task framework with two baselines as shown in Table~\ref{result-pr}. For passage selection, our multi-task model achieves the accuracy of 38.9, which outperforms the pure answer prediction model with 4.3. Moreover, jointly learning the answer prediction part and the passage ranking part is better than solving this task by two separated steps because the answer span can provide more information with stronger supervision, which benefits the passage ranking part. The ROUGE-L is calculated by the best answer span in the selected passage, which shows our multi-task learning framework has more potential for better answer.
		
		\subsubsection{Extraction vs. Synthesis}
		We compare the result of answer extraction and answer synthesis in different categories grouped by the upper bound of extraction method in Table~\ref{marco-compare}. For the question whose answer can be exactly matched in the passage, our answer synthesis model performs slightly worse because the sequence-to-sequence model makes some deviation when copying extracted evidences. In other categories, our synthesis model achieves more or less improvement. For the question whose answer can be almost found in the passage (ROUGE-L$\geq$0.8), our model achieves 0.2 improvement even though the space that can be raised is limited. For the question whose upper performance via answer extraction is between 0.6 and 0.8, our model achieves a large improvement of 2.0. Part of questions in the last category (ROUGE-L$<$0.2) are the polar questions whose answers are ``yes'' or ``no''. Although the answer is not in the passage or question, our synthesis model can easily solve this problem and determine the correct answer through the extracted evidences, which leads to such improvement in this category. However, in these questions, answers are too short to influence the final score in terms of BLEU because it is normalized in all questions. Moreover, the score decreases due to the penalty of length. Due to the limitation of BLEU, we only report the result in terms of ROUGE-L in our analysis.
		
		\begin{table}
			\centering
			\begin{tabular}{lcc}
				\hline
				\bf Method & \bf P@1 & \bf ROUGE-L \\
				\hline
				Extraction w/o Passage Ranking & 34.6 & 56.7 \\
				Passage Ranking then Extraction & 28.3 & 52.9\\
				S-Net (Extraction) & \bf 38.9 & \bf 59.4 \\
				\hline
			\end{tabular}
			\caption{Results of passage ranking. -w/o Passage Ranking: the model that only has evidence extraction part, without passage ranking part. -Passage Ranking then Extraction: the model that selects the passage firstly and then apply the extraction model only on the selected passage. }\label{result-pr}
		\end{table}
		
		\begin{table}
			\begin{center}
				\begin{tabular}{lcc}
					\hline
					\multirow{2}{*}{\bf Category} & \bf 	\multirow{2}{*}{\bf Extraction} & \bf Extraction \\
					& & \bf +Synthesis \\
					\hline
					max = 1.0 (63.95\%) & 50.74 & 49.59 \\
					0.8$\leq$max$<$1.0 (20.06\%) & 40.95 & 41.16 \\
					0.6$\leq$max$<$0.8 (5.78\%)& 31.21 & 33.21 \\
					0.4$\leq$max$<$0.6 (1.54\%) & 21.97 & 22.44 \\
					0.2$\leq$max$<$0.4 (0.29\%) & 13.47 & 13.49 \\
					max$<$0.2 (8.38\%) & 0.01 & 49.18 \\
					\hline
				\end{tabular}
			\end{center}
			\caption{The performance of questions in different levels of necessary of synthesis in terms of ROUGE-L on MS-MARCO development set.}
			\label{marco-compare}
		\end{table}
		
		\subsubsection{Comparison with the End-to-End Generation Framework}
		We compare our extraction-then-synthesis model with several end-to-end generation models in Table \ref{end-to-end}. 	
		S2S represents the sequence-to-sequence framework shown in Figure~\ref{s2s-network_overview}. The difference among our synthesis model and all entries in the Table~\ref{end-to-end} is the information we use in the encoding part. The authors of MS-MACRO publish a baseline of training a sequence-to-sequence model with the question and answer, which only achieves 8.9 in terms of ROUGE-L. Adding all passages to the sequence-to-sequence model can obviously improve the result to 28.75. Then we only use the question and the selected passage to generate the answer. The only difference with our synthesis model is that we add the position features to the basic sequence-to-sequence model. The result is still worse than our synthesis model with a large margin, which shows the matching between question and passage is very important for generating answer. 	
		Next, we build an end-to-end framework combining matching and generation. We apply the sequence-to-sequence model on top of the matching information by taking question sensitive passage representation $v^P_t$ in the Equation \ref{vtp} as the input of sequence-to-sequence model, which only achieves 6.28 in terms of ROUGE-L.			
		Above results show the effectiveness of our model that solves this task with two steps. In the future, we hope the reinforcement learning can help the connection between evidence extraction and answer synthesis.	
		
		\begin{table}
			\begin{center}
				\begin{tabular}{lc}
					\hline
					\bf Method &\bf ROUGE-L \\ 
					\hline
					S2S (Question) & 8.9\\
					S2S (Question + All Passages) & 28.75 \\
					S2S (Question + Selected Passage) & 37.70 \\
					\hline
					Matching + S2S & 6.28 \\
					\hline 
					
				\end{tabular}
			\end{center}
			\caption{The performance on MS-MARCO development set of end-to-end methods.}
			\label{end-to-end}
		\end{table}

		\section{Conclusion and Future Work}
		In this paper, we propose S-Net, an extraction-then-synthesis framework, for machine reading comprehension. The extraction model aims to match the question and passage and predict most important sub-spans in the passage related to the question as evidence. Then, the synthesis model synthesizes the question information and the evidence snippet to generate the final answer. We propose a multi-task learning framework to improve the evidence extraction model by passage ranking to extract the evidence snippet, and use the sequence-to-sequence model for answer synthesis. We conduct experiments on the MS-MARCO dataset. Results demonstrate that our approach outperforms pure answer extraction model and other existing methods.
		
		We only annotate one evidence snippet in the sequence-to-sequence model for synthesizing answer, which cannot solve the question whose answer comes from multiple evidences, such as the second example in Table \ref{example}. Our extraction model is based on the pointer network which selects the evidence by predicting the start and end positions of the text span. Therefore the top candidates are similar as they usually share the same start or end positions. By ranking separated candidates for predicting evidence snippets, we can annotate multiple evidence snippets as features in the sequence-to-sequence model for questions in this category in the future.

	\section*{Acknowledgement}
	We thank the MS-MARCO organizers for help in submissions.

	\bibliography{iclr2017_conference}
	\bibliographystyle{iclr2017_conference}
	
\end{document}